\documentclass[letterpaper, 10 pt, conference]{ieeeconf}  

\IEEEoverridecommandlockouts                              

\usepackage{graphics} 
\usepackage{amsmath} 
\usepackage{todonotes}
\usepackage{comment}
\usepackage{diagbox}
\usepackage{makecell}

\title{\LARGE \bf
Proprioception and reaction for walking among entanglements
}

\author{Justin K. Yim$^{1,2}$, Jiming Ren$^{2}$, David Ologan$^{2}$, Selvin Garcia Gonzalez$^{2}$, and Aaron M. Johnson$^{2}$
\thanks{This work was supported in part by the National Science Foundation under Grant CCF-2030859 to the Computing Research Association for the CIFellows Project, Grant ECCS-1924723, the GEM Consortium Fellowship, and by funding award \#HQ00342110020 from the National Defense Education Program.}
\thanks{
$^{1}$Department of Mechanical Science and Engineering, University of Illinois Urbana-Champaign, Urbana, IL 61801, USA {\tt\small jkyim@cmu.edu}}%
\thanks{
$^{2}$Department of Mechanical Engineering, Carnegie Mellon University, 
Pittsburgh, PA 15232, USA, {\tt\small amj1@cmu.edu}}%
}

\begin{document}
\bstctlcite{IEEEexample:BSTcontrol} 

\maketitle
\thispagestyle{empty}
\pagestyle{empty}

\begin{abstract}
Entanglements like vines and branches in natural settings or cords and pipes in human spaces prevent mobile robots from accessing many environments.
Legged robots should be effective in these settings, and more so than wheeled or tracked platforms, but naive controllers quickly become entangled and stuck.
In this paper we present a method for proprioception aimed specifically at the task of sensing entanglements of a robot's legs as well as a reaction strategy to disentangle legs during their swing phase as they advance to their next foothold.
We demonstrate our proprioception and reaction strategy enables traversal of entanglements of many stiffnesses and geometries succeeding in 14 out of 16 trials in laboratory tests, as well as a natural outdoor environment.
\end{abstract}

\section{INTRODUCTION}

Tripping hazards like vines, branches, and outcroppings fill many natural environments from forest floors to reed beds.
Without a way to navigate these entanglement-filled environments, mobile robots cannot perform important tasks like environmental monitoring, scientific sampling, or firefighting.
In human environments, robots may need to navigate around cords, hoses, and protrusions to perform tasks like inspection, delivery, or in-home assistance.
In this work, we develop proprioceptive detection of entanglements on a quadruped robot's swing legs and simple control strategies to disentangle from them, enabling the robot to walk through numerous, distributed contacts of any stiffness.

In highly vegetated settings, legged robots show unique promise compared to alternative robotic platforms.
Small wheels become stuck in vines and branches.
Large wheels and tracks roll over obstacles, but may crush or damage them in the process.
Flying robots' large wings, bodies, and rotors do not fit between dense woody plants.
Legged robots have the flexibility to both choose footholds and adjust posture to disentangle from and move around obstructions.
However, as we show in the experimental section, standard controllers that do not reason about motion through entanglements rapidly become stuck on vine-like obstacles.

Moving through obstacles like thickets or reeds, contact is unavoidable.
A locomotor must press through them to progress and cannot simply avoid contact with anything a visual sensor detects is solid.
A locomotor should be able to push through flexible obstructions, but detect when it is stuck so that it can disentangle itself.
However, the stiffness of obstacles is difficult to gauge prior to contact.
Foliage blocks most visual sensors and the obstructed materials behind can vary in stiffness from flexible twigs to rigid rocks.

Detection is also complicated by the dynamics of legged locomotion.
Feet make and break contact with every stride.
During stance phases, forces on the feet are a large fraction of bodyweight.
However, during swing phases, forces much smaller than bodyweight can be disruptive if they are applied at unexpected contacts on the body, like tripping over a foot caught on a ledge or knocking over a chair bumped by a knee.
Rapidly detecting these unexpected forces on the legs is particularly challenging since they are interspersed between the stance phase forces and smaller in magnitude than them.

\begin{figure}
    \centering
    \includegraphics[width=1.0\linewidth]{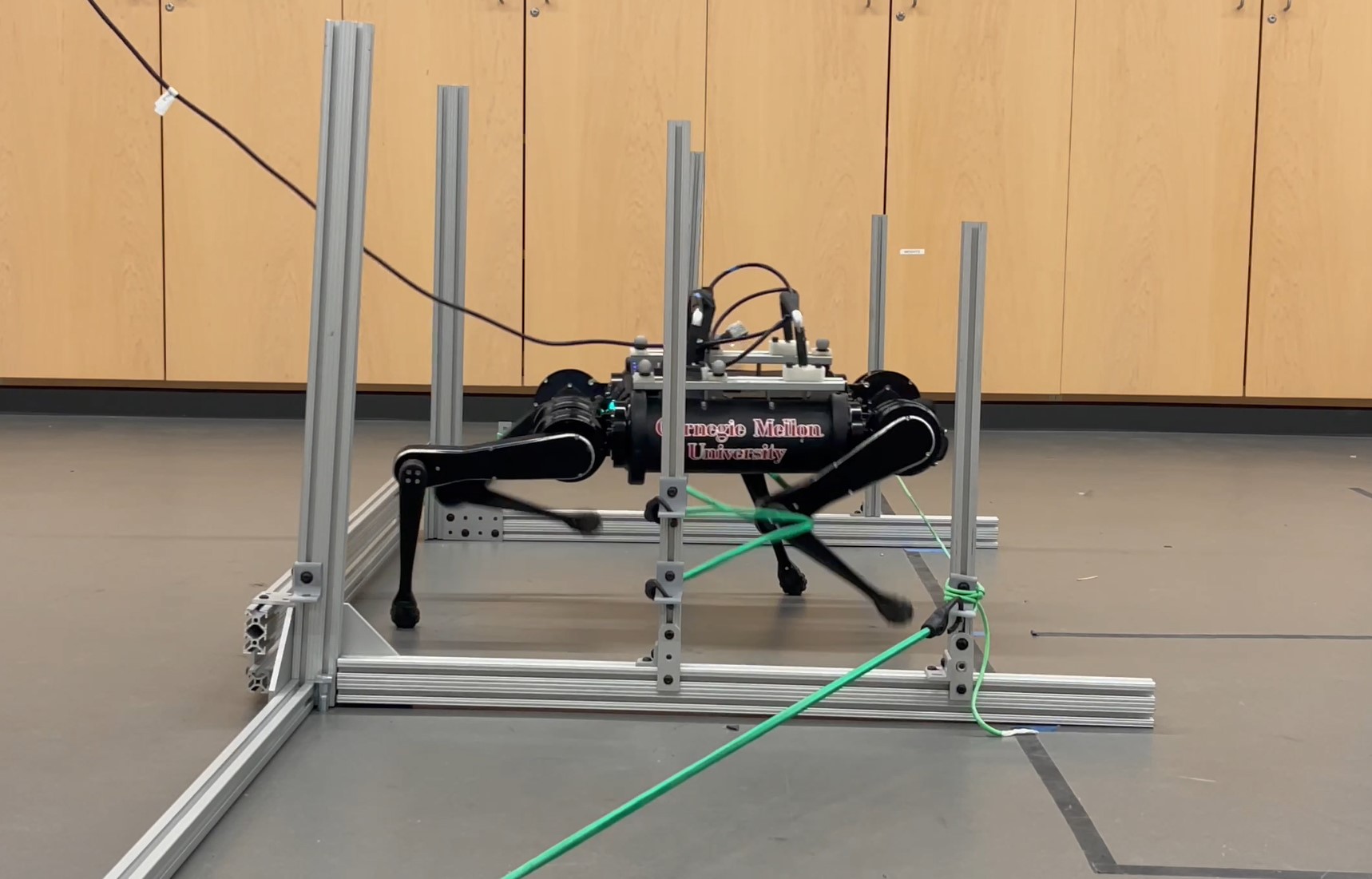}
    \caption{This paper presents an approach to enable a quadruped robot to walk through entanglements like the bar, bungee cord, and rope shown here.}
    \label{fig:intro}
    \vspace{-0.15in}
\end{figure}

Prior work has demonstrated many approaches to estimating contact.
An excellent overview can be found in \cite{Haddadin2017}.
Often, proprioception enables robot arms to detect collisions with objects and people (as discussed in \cite{Haddadin2017}) including when mounted on a legged robot \cite{VanDam2022}.
\cite{Bhatia2022} analyzes the timing and impulse associated with contact detection with rigid objects.
However, robot arms are not usually subjected to the large recurring forces experienced by legs.

On legs, proprioception often provides foot contact detection \cite{Bledt2018} but reasoning about contacts at other locations on legs is not as common.
\cite{Wang2020} localizes where contacts are on legs or fingers but assumes rigid objects.
Measuring non-stance forces on the legs enables adaptation to these disturbances \cite{Morlando2021} and obstacle avoidance \cite{Johnson2010}.

Other work adds additional sensors for detecting contact.
Some use vision to address obstacles or combine vision with proprioception.
Lidar and radar can detect obstacles even behind foliage, but processing is significant and obstacle stiffness is still unknown \cite{Matthies2005}.
Proprioception can augment vision to detect obstacles or terrain parameters like friction not observed by visual sensors \cite{Fu2022} or estimate ground surface height when it is occluded by foliage \cite{Homberger2019}.
Alternatively, adding sensing skins allows a robot to measure contact with great granularity.
However, skins involve significant additional hardware, complexity, and potential for damage \cite{Ohmura2007,Cannata2008,Buchan2013}, or may not be able to measure many simultaneous contacts from plants or other objects \cite{Fan2022}.

Sensing disturbances like unexpected contact by any of the discussed methods is very useful since it enables a robot to trigger reactions to them.
Stepping over rigid obstacles like curbs can avert tripping, shown in simulation by \cite{Boone1997} and hardware by \cite{Focchi2013}.
We aim to extend simple reactions to address obstacles of variable geometry and compliance including thin entangling objects like vines and cords.

We present a proprioceptive reactive approach that enables a quadruped robot to walk through entanglements distributed over its limbs.
This contribution consists of two key components.
First, we modify momentum-based observation informed by the hybrid forces of walking to better detect lower-magnitude forces on a leg in swing phase between its stance phases, described in Section~\ref{sec:MBO} and hardware tested in Section~\ref{sec:MBO results}.
Second, we develop a reactive swing-leg disentanglement controller that enables a robot to slide out of and step over simple entanglements described in Section~\ref{sec:reaction} and hardware tested in Sections~\ref{sec:baselines} through \ref{sec:outdoors}.

\section{METHODS}

\begin{figure}
    \centering
    \vspace{.5em}
    \includegraphics[width=1.0\linewidth]{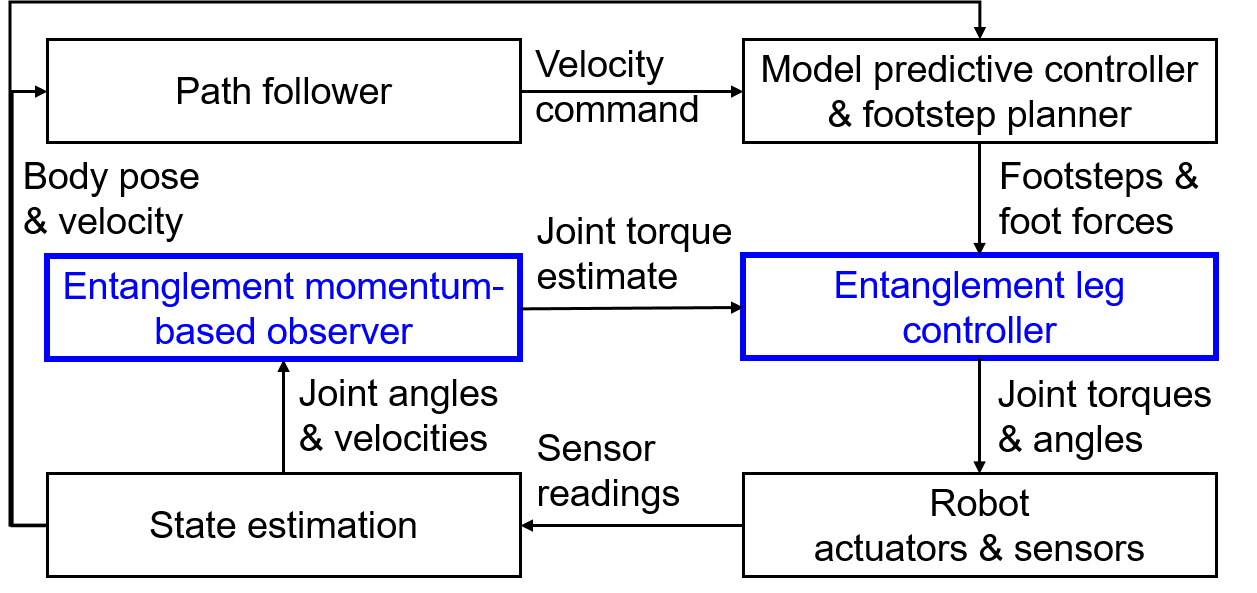}
    \caption{High-level block diagram of the proposed approach integrated into a planning and controls stack with new contributed modules highlighted in blue.}
    \label{fig:block diagram}
    \vspace{-0.15in}
\end{figure}

Our methods consists of two control modules: a momentum-based observer suited for detecting swing-leg entanglements, and a swing-leg controller (Fig.~\ref{fig:block diagram}). It is implemented as an extension to the Quad-SDK open-source software \cite{Norby2022} for legged robot control.

\subsection{Momentum-based Observer}
\label{sec:MBO}

\begin{figure}
    \centering
    \includegraphics[width=0.75\linewidth]{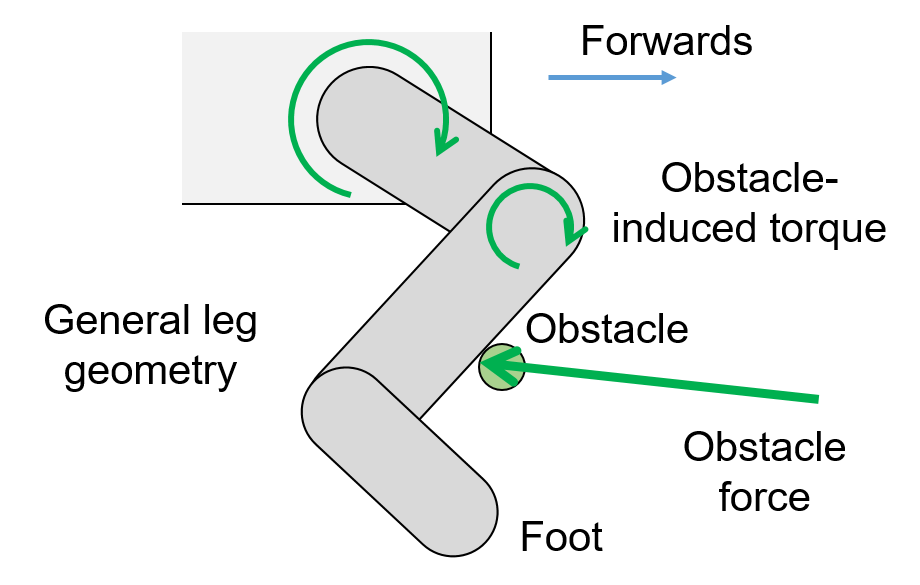}
    \caption{General leg geometry for obstacle-induced torque estimation}
    \label{fig:legMO}
    \vspace{-0.15in}
\end{figure}

In this section, we develop proprioceptive estimation that uses only existing sensors on a commercial quadruped robot (motor encoders and current monitoring) to provide feedback about contacts across the limbs that may impede progress.
We adapt the Momentum-Based Observer (MBO) of \cite{Haddadin2017} to monitor externally-induced joint torques on a quadruped robot's legs.
A new innovation is the re-initialization of the MBO to improve detection of small forces between stance phases as described in section~\ref{sec:initialization}.

We assume that the robot attempts to walk forwards through entanglements at a steady speed with level torso orientation and height and that the torso has significantly more inertia than the limbs.
Following these assumptions, we use the robot's torso frame as the base inertial reference frame, ignoring its acceleration and rotation.
Experiments in Section~\ref{sec:results} show that these assumptions hold.

We model the robot's limbs as four independent serial chains with equations of motion governed by
\begin{align}
    M(q) \ddot{q} + C(q, \dot{q}) \dot{q} + G(q) - \tau_m - \tau_f =  \tau_\text{ext}
    \label{eq:EOM}
\end{align}
where $q$ are the joint angles, $M(q)$ is the inertia matrix, $C(q, \dot{q})$ contains the Christoffel symbols, $G(q)$ are the gravitational terms, $\tau_m$ are the joint torques applied by the motors, $\tau_f$ are estimated joint friction torques, and $\tau_\text{ext}$ are the externally-induced joint torques.
For the Spirit-40 quadruped robot, each leg has three joints (named ab/ad, hip, and knee from proximal to distal and rotating about the body $x$, $y$, and $y$ axes in the zero configuration), such that $q$, $\tau_m$, and $\tau_\text{ext}$ are vectors of length three.
$\tau_\text{ext}$ arises from the sum of all external wrenches applied to the limb
\begin{align}
    \tau_\text{ext} = \sum_{i=0}^n J_{c_i}^T(q) F_{\text{ext}, i}
\end{align}
where $F_{\text{ext}, i}$ is one of the $n$ wrenches applied at point $c_i$ on the limb and $J_{c_i}(q)$ is the geometric contact Jacobian. 

Taking the time derivative of the generalized momentum $p = M(q) \dot{q}$ and solving for $\tau_\text{ext}$ produces the momentum-based observer described in \cite{Haddadin2017} section III-E, specifically equations (42--44).
The resulting observer maintains a linear first-order estimate $r$ that approximates $\tau_\text{ext}$ and does not require noisy double differentiation of joint angles $q$
\begin{align}
    \hat{\beta}(q,\dot{q}) &= G(q) - C^T(q,\dot{q})\dot{q} \\
    \dot{\hat{p}} &= \tau_m - \hat{\beta}(q,\dot{q}) + r \\
    r &= K_O (p - \hat{p})
\end{align}
where $\hat{p}$ is the estimated generalized momentum of the limb and $p$ is the generalized momentum of the limb computed from sensor measurements.
Note that we have simplified the dynamics of $r$ compared to \cite{Haddadin2017}.
$K_O$ is the diagonal observer gain matrix that can be selected to tune the time constant of the observer trading off response time and smoothing of noise.
We chose all diagonal elements of $K_O$ to be 25 Hz tuned empirically for a balance between fast response and rejection of high-frequency noise.
We identified inertial parameters for computing $M$, $C$, and $G$ using the method of \cite{Lee2020} with a slight modification to also estimate friction parameters $c$ and $d$ which we used for friction compensation:
\begin{align*}
    \tau_{f,i} = -c \, \text{sign}(\dot{q}_i) - d \dot{q}_i
\end{align*}
We found that joints of the same type exhibited similar friction parameters with the highest dry friction of $c = 0.44$ N m in the knees and that viscous damping was small for all joints.

\subsubsection{Torque estimates for reaction}
The MBO torque estimates do not distinguish the various forces or points of application on the robot's limb, but provide an estimate of the total resistance the robot feels and its direction.
Solving the full location and magnitude of every contact distributed across the limb is an under-specified problem \cite{Wang2020} but the full solution is not required to react to obstacles that impede the robot's motion; the MBO's joint torque estimates suffice.

Additional simplification arises from the task of moving forwards in one direction and a few realistic assumptions about limb morphology and contact forces.
First, we assume that each of the robot's legs is an unbranching serial chain of broadly straight and smooth links in which each link extends only below and not above the preceding joint as is usual for legs.
Second, we assume that obstacles cannot exert pulling forces on the surface of the robot limb -- that is, there are negligible adhesive forces.
Given these assumptions, only obstacles ahead of the robot limb can exert backwards-directed forces impeding forwards progress.
Furthermore, backwards-directed forces always exert torques in only one direction on joints with axes that are horizontal and perpendicular to the direction of forwards motion (as in humans' knees and in Spirit 40's hip and knee joints).
This allows us to ignore torques in the opposite direction, since they can arise only from obstacles pressing on the rear surface of the limb where they will not impede progress.

\subsubsection{Initialization}
\label{sec:initialization}
An interesting challenge for external contact force estimation arises from the action of walking in which intentional forces with the world alternate between large (in stance) and small (in swing) magnitudes.
Thus, in this paper we extend the momentum-based observer approach to separate the estimation of smaller forces from the planned large external forces.

During stance, external force is usually on the order of 1/2 of the bodyweight for a walking trot gait at 50\% duty factor, resulting in two legs supporting the body at any moment.
However, even forces that induce smaller torques at the joints can cause an impediment to swing-leg motion.

The robot's maximum pushing force is limited, requiring the MBO to detect forces smaller than those at which the robot becomes stuck.
Forces applied to the swing legs totalling more than some force $F_\text{max}$ are too large for the legs to push through during swing.
At the upper bound, $F_\text{max}$ can be no larger than $\mu m g$ where $\mu$ is the coefficient of friction and $m g$ is the bodyweight of the robot.
However, in reality this limit is lower due to controller performance and avoiding toppling the robot.
The achieved $F_\text{max}$ depends on the robot's forward velocity control gain tuning, commanded velocity, and terrain slope and friction.

$F_\text{max}$ sets an upper limit on swing forces, while the MBO's noise floor provides a minimum force threshold for detection.
Its accuracy depends on inertial parameter identification errors whose effect grows as joint velocity and acceleration increase and on joint dry friction that makes it difficult to measure torques lower in magnitude than the friction force.

Detecting swing-phase contacts early is important  for there to be sufficient time for any reaction to clear the obstacle.
However, soft contacts from vines or brush in swing phase can be more difficult than traditional hard impacts since no impulsive rapid change in momentum results.
Furthermore, these smaller forces are easily overshadowed by the large stance phase forces early in swing as the momentum observer estimates converge from their initial large stance values.

To rapidly detect forces on the legs, we re-zero the momentum-based observer's momentum estimate $\hat{p}$ as the leg leaves stance phase and begins its swing motion.
In physical implementation, the re-zeroing period lasts 30 ms at the start of swing in case stance ends late due to state uncertainty or communication latency.
This reinitialization takes advantage of knowledge about stance and swing behavior to improve accuracy at the start of swing since the limb begins close to rest and nominally under little load as it ends stance phase and begins swing phase.
We show the benefit of the proposed MBO reinitialization in Section~\ref{sec:MBO results}.

\subsection{Entanglement reaction}
\label{sec:reaction}

\begin{figure*}
    \centering
    \vspace{.25em}
    \includegraphics[width=1.0\textwidth]{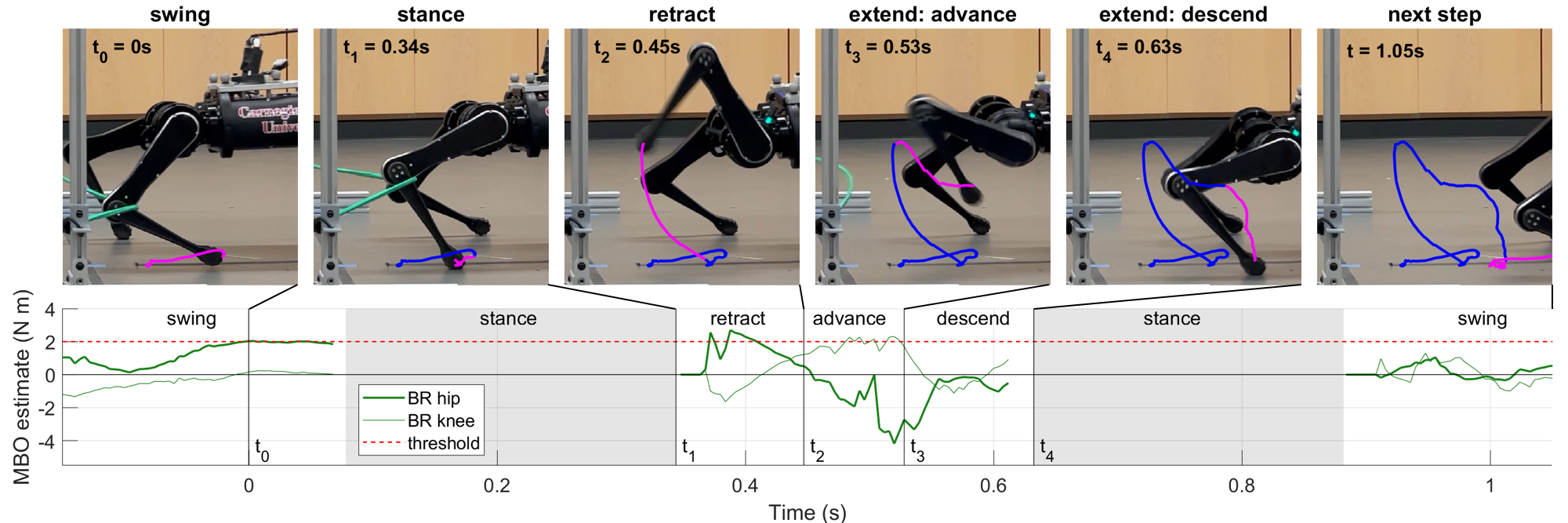}
    \caption{Spirit 40 detects and disentangles from a bungee cord.
    The back right leg detects entanglement when the hip torque exceeds the threshold near the end of a swing phase ($t_0 = 0$s).
    During the next swing phase it disentangles via three motions: retract ($t_1$--$t_2$), extend advance ($t_2$--$t_3$), and extend descend ($t_3$--$t_4$).
    Following disentanglement, the leg reverts to default behavior on the following step.
    Pink trace indicates the motion of the foot since the previous frame, while blue shows all previous motion of the foot.}
    \label{fig:reaction}
    \vspace{-0.15in}
\end{figure*}

\subsubsection{Formulation and gait constraints}
\label{sec:formulationandgait}
Conventional controllers walk well in the absence of obstacles, but a change in gait or ``reaction" may be required if a limb makes contact with an obstacle.
Two important obstacle properties are: the peak horizontal force $F_{o}$ with which it can oppose forward motion of a limb; and the notion of its geometric ``complexity" or how difficult it would be for a path planner to find a feasible trajectory to overcome the obstacle.
Reactions can occur at different levels of the controller with phase transitions occuring based on these parameters:
\begin{enumerate}
    \item If $F_{o}$ is below $F_\text{max}$, gait kinematics may remain largely unchanged and the swing leg can push through the obstacle by using higher swing leg force.
    \item If $F_{o}$ is greater than $F_\text{max}$ but the entanglement geometry is not complex, the swing leg can be retracted to step over the obstacle.
    \item If the entanglement geometry is moderately complex, the robot can adjust its gait to solve the entanglement.
    \item If the entanglement geometry is exceedingly complex, the robot can give up on its current planned path and replan a path that goes elsewhere.
\end{enumerate}

Each method corresponds to reaction at a different level in a conventional hierarchical planning and control framework escalating up from joint control to path planning.
Generally, higher-level reactions involve more aggressive deviation from nominal behavior with corresponding slower progress and greater energetic cost.

In this work we focus on level 2) where the robot can retain a high speed and address moderately complex terrain that may be encountered frequently in realistic environments.
Adaptation of the swing leg motion involves two steps: ``retraction" and ``extension" outlined below.

\subsubsection{Leg retraction}
\label{sec:legretraction}
Leg retraction during swing phase aims to slide out of contact with any obstacles in front of the leg without precise knowledge of the object's geometry or material properties like stiffness.
Once $\tau_\text{ext}$ at the hip or knee exceeds an experimentally chosen threshold (2 Nm for Spirit 40 in only the positive direction, as described in Section~\ref{sec:MBO}) during swing, the leg is considered ``stuck" and begins retraction.
Ideally, leg retraction should circumvent any number of obstacles of any stiffness contacting the leg at any point on its front surface.
Retraction has three objectives:
\begin{enumerate}
    \item Raise the bottom point of the leg over obstacles
    \item Slide obstacles off the bottom of the leg
    \item Retain contact until obstacles slide off the bottom of the leg to avoid retreating from and advancing into the same obstacle in the absence of visual feedback.
\end{enumerate}
Essentially, while torque estimates indicate the leg is still behind an obstacle, the leg should apply an upward velocity at the foot and a light forwards pressure with its distal links.
This is akin to hybrid force-velocity control in which the foot of the robot applies a forwards force to retain contact but achieves a positive vertical velocity to slide above them.
In order to slide contacts off of the bottom of the leg, front surfaces of the limb should retain a downward slope.

For the simple 2-link leg of a commercial quadruped robot, an experimentally tuned strategy can be applied using a retraction velocity command at the hip joint (15 rad/s) and a forwards torque command (2 Nm) at the knee joint.
In physical implementation, retraction lasts a minimum of 0.1 seconds to ensure obstacles close to the knee are cleared.

\paragraph{Extension}
Once no more contacts are detected on the limb ($\tau_\text{ext}$ drops below 2 Nm for Spirit 40) the limb is considered ``free" again and can extend towards the desired foothold.
Extension also begins if the end of the allotted swing phase time is soon, less than parameter $t_{down}$ s away, to ensure it is on the ground in time for the next stance.
In order to avoid re-entangling on obstacles, the lowest point on the leg (the foot) remains high subject to kinematic constraints and ``advances" horizontally until the foot is above the desired foothold.
At this point, the foot ``descends" until it contacts the ground.
By first advancing and then descending, any obstacle between the previous foothold and the next foothold (below the height at which obstructions slid off) will either not contact the leg or will contact only rear surfaces of the leg.
One full reaction motion to disentangle from a bungee cord is illustrated in Fig.~\ref{fig:reaction}.

\subsection{Gait}

Gait period $T$ is a particularly important parameter since it impacts walking control stability, MBO accuracy, and reaction motion success.
Low step frequency reduces stability.
In experiments walking at a commanded 0.5 m/s, the regularity of the gait and error from the commanded posture, height, and speed deviated further as the gait period increased.
Spirit 40 regulated its forward velocity with a 0.037 m/s standard deviation with $T = 0.36$ s, rising to 0.084 m/s at $T = 0.54$ s and to 0.198 m/s at $T = 0.75$ s.

However, shorter values of $T$ necessitate faster reaction motion to clear high obstacles.
Swinging leg joints through $\pi/2$ rad and back during one step in a 50\% duty factor trot requires an average joint speed of $2 \pi/T$.
With Spirit 40's knee joint free running speed of approximately 20 rad/s, $T$ can be no shorter than 0.31 s and in practice should be somewhat slower to leave sufficient control bandwidth.
As a consequence, we standardized on a gait period of 0.54 s, 50\% slower than the Quad-SDK default of 0.36 s.

The robot often detects entanglement too late to complete a successful retraction; e.g.\ if the leg hits the obstacle late in its swing.
In these cases, the leg can begin its next swing already retracting to provide ample time to clear the obstacle.
To achieve this, the stuck/free state at the beginning of the next swing is retained from the stuck/free state at time $t_{down}$.
If a component of $\tau_\text{ext}$ remains high at time $t_{down}$, the next swing begins retracting while if the leg is free at $t_{down}$ the next swing begins by advancing towards the foothold.

\subsection{Experimental Setup}
\begin{figure}
    \centering
    \vspace{.25em}
    \includegraphics[width=1.0\linewidth]{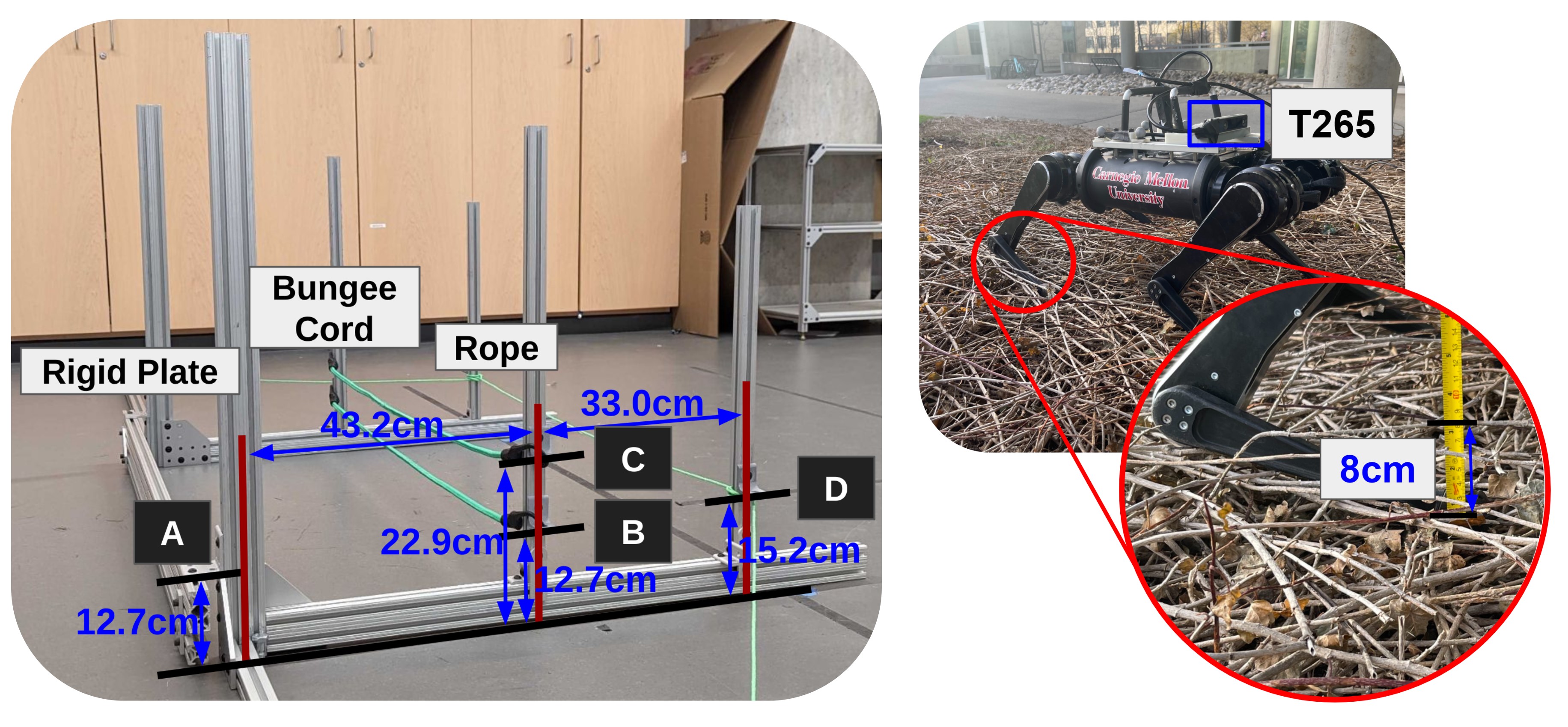}
    \caption{(Left) Indoor MOCAP testing setup with four anchor points for obstacles. The image shows the mixed stiffness configuration: a rigid plate at anchor A, two bungee cords at anchors B and C, and a rope at anchor D. (Right) Outdoor testing setup in natural entanglement.}
    \label{fig:indoor setup}
    \vspace{-0.15in}
\end{figure}

\subsubsection{Indoor Experiments}
We test the baseline and swing retraction walking strategies on hardware using Ghost Robotics' Spirit-40 to step over obstacles with various stiffnesses and altitudes. Our indoor testing uses motion capture (MOCAP) to provide an estimate of body position and orientation for the robot. We present the result of using different walking strategies when facing baseline soft bungee cords in Table \ref{tab:baselines}, and the result of applying our method to go over five different arrangements of obstacles in Table \ref{tab:obstacles}. Our obstacle setup considers Spirit-40's nominal stance CoM height of 27 cm and is shown in Fig.~\ref{fig:indoor setup} (Left). We tie four light-duty bungee cords (rest-length 1.2 m and a stiffness of about 250 N/m for deflections under 8 cm, softening to about 70 N/m for larger deflections) on all anchors \emph{A} to \emph{D}. The rope and rigid plate are placed at anchor \emph{A} when tested in isolation. In the mixed setting, we put one rigid plate on anchor \emph{A}, two bungee cords on anchors \emph{B} and \emph{C} and one rope on anchor \emph{D}. In the net experiments, a soccer net with 14 cm square openings covers the entire aisle at an average height of 7.6 cm. In all settings, the robot is programmed to go straight at a constant speed of 0.5 m/s with a gait period of 0.54 s unless otherwise noted.

\subsubsection{Outdoor Setup}
In the outdoor setting, we deploy Spirit-40 in interwoven natural vines shown in Fig.~ \ref{fig:indoor setup} (Right).
The vines pile up approximately 8 cm above the ground.
An onboard Intel Realsense T265 camera provides rough body pose estimates to replace indoor tests' MOCAP estimation.
The robot walking speed is decreased to 0.4 m/s for more stable estimation. 

\section{RESULTS}
\label{sec:results}

We tested our momentum-based observer and disentanglement reaction in a series of hardware experiments.
First, we tested that the MBO sensitivity enables distinguishing between obstacles that can be pushed past and those that must be avoided.
Second, we compared our strategy to simple baseline controllers and demonstrated that our strategy succeeds where others fail while using less power than aggressive motion.
Third, we tested our strategy across a variety of different obstacle parameters including not just height but also stiffness.
Finally, we tested our strategy against many complicated contacts in and out of the lab.

\subsection{Momentum-based Observer}
\label{sec:MBO results}

\begin{figure}
    \centering
    \includegraphics[width=1.0\linewidth]{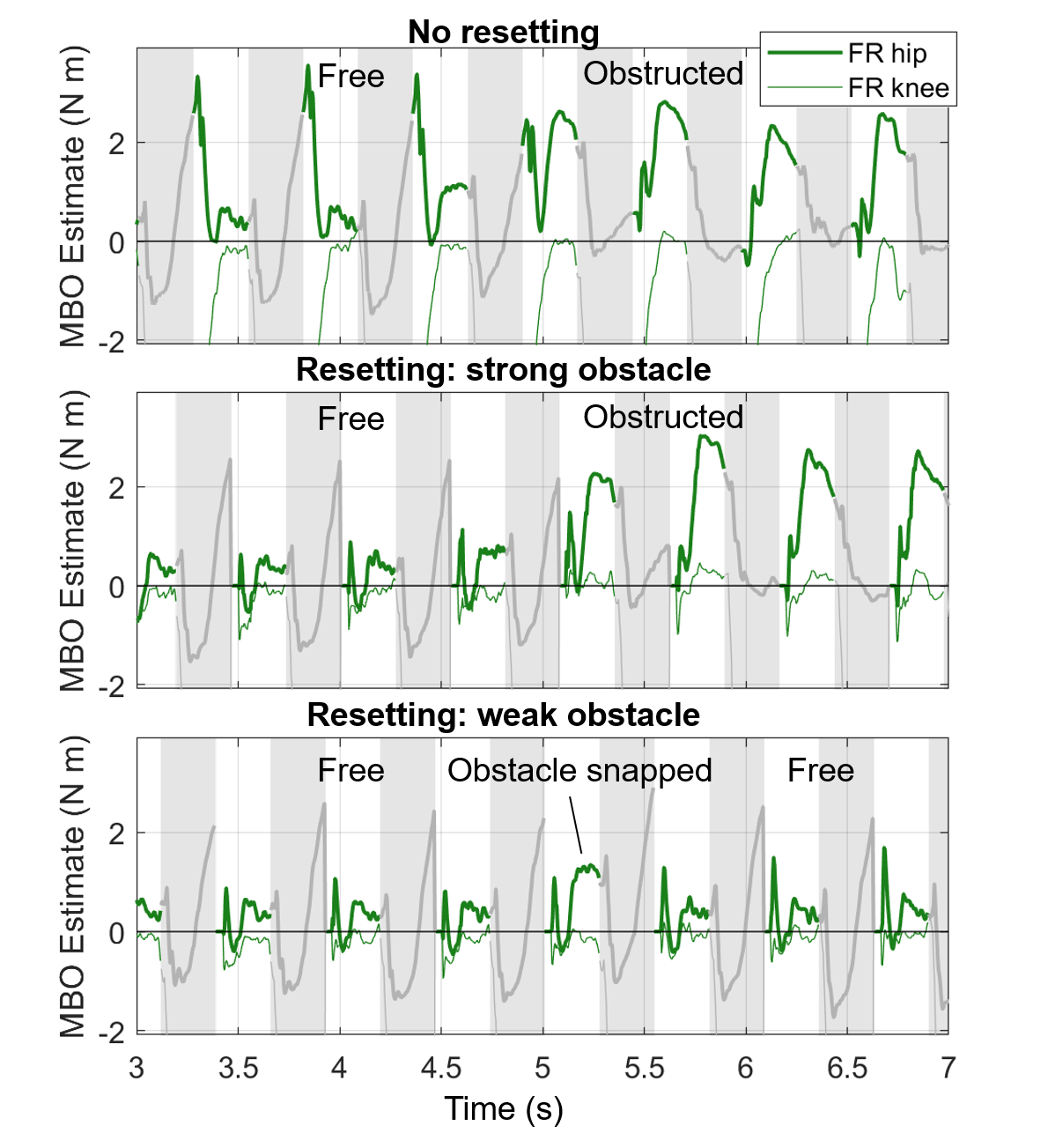}
    \caption{(Top) Without resetting, the Momentum-Based Observer (MBO) retains high torque estimates from stance at the beginning of swing even when no contacts are present (early), potentially masking actual contacts on the swing leg (later).
    (Middle) Resetting the MBO estimates to zero at the start of stance is a more accurate initialization, allowing easy thresholding to distinguish free (early) and obstructed (later) swing motion.
    (Bottom) Weak obstacles like a thin wire that is easily broken register as small forces.}
    \label{fig:MO_reset}
\end{figure}

We tested the momentum-based observer walking with and without resetting (Fig.~\ref{fig:MO_reset} top and middle) and into obstacles of different strengths -- a bungee cord and an easily broken wire (Fig.~\ref{fig:MO_reset} middle and bottom).

By resetting the MBO at the beginning of swing, large torques from stance phase do not corrupt the initial swing phase torque estimate, allowing it to easily distinguish between negligible and significant forces by a simple threshold on the torque magnitude.
As stated earlier, forces pushing on the front surfaces of the limbs can exert only positive torques about the joints so the threshold needs to only be applied in the positive direction for all joints in the leg.

\subsection{Reactionless strategy baselines}
\label{sec:baselines}

\begin{figure}
    \centering
    \vspace{.25em}
    \includegraphics[width=1.0\linewidth]{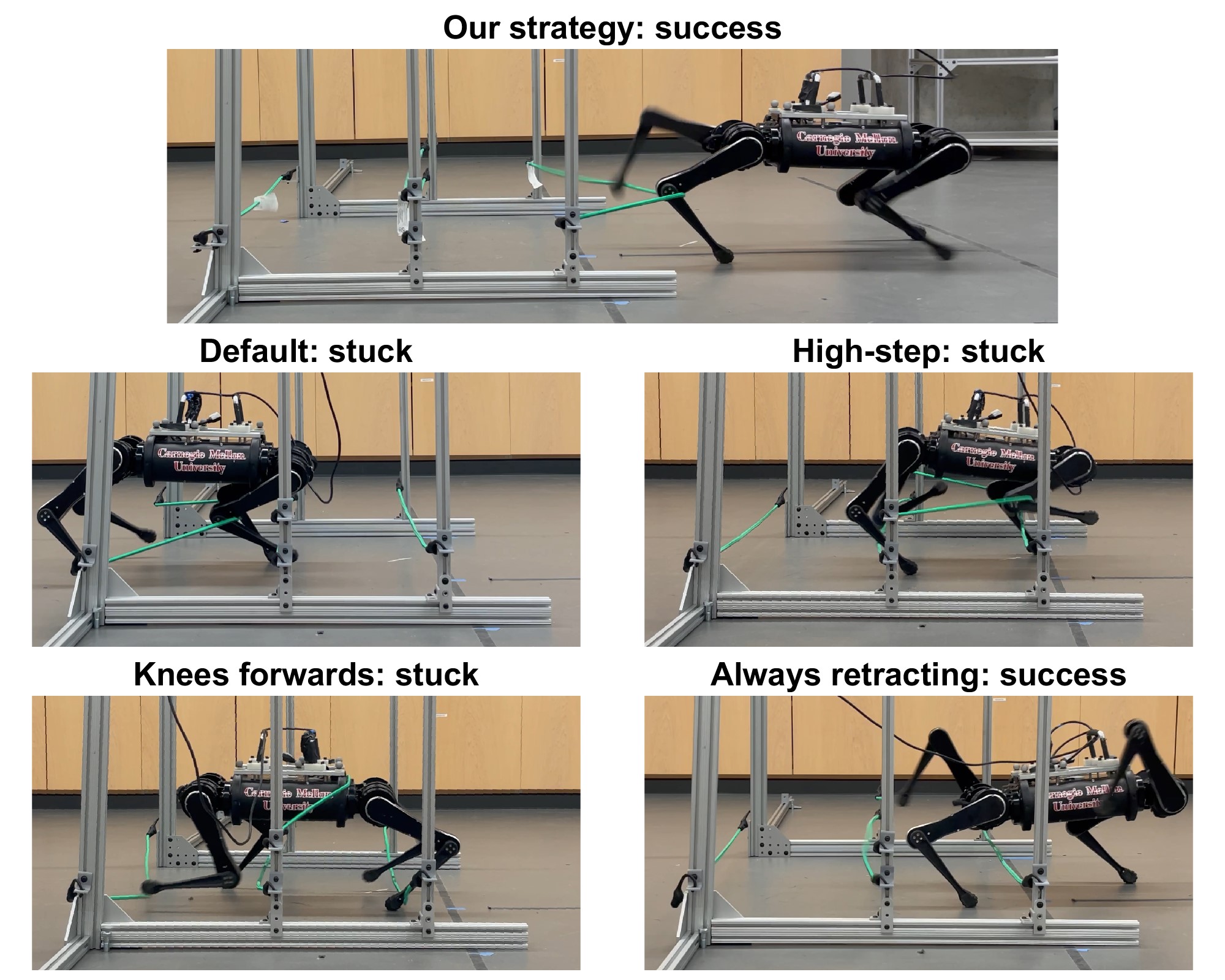}
    \caption{Our strategy and non-reactive baseline strategies walk into four elastic obstacles: our strategy overcomes obstacles that stymied baselines 1--3 while using less energy than baseline 4.}
    \label{fig:baseline snapshots}
\end{figure}

\begin{table}
    \begin{tabular}{l|rrrrll}
        \backslashbox{Strat.}{Metric} & \makecell{Success\\trials} & \hspace{-.5em}\makecell{Obstacles \\ cleared} & \makecell{$P_f$ \\ W} & \makecell{$P_o$ \\ W} & \makecell{$v_f$ \\ m/s} & \makecell{$v_o$ \\ m/s} \\
        \hline
        Default \cite{Norby2022} & 0/3 & 0/12 & 87 & 133 & 0.49 & 0\\
        High step & 0/3 & 2/12 & 126 & 191 & 0.46 & 0\\
        Knee forward & 0/3 & 6/12 & 94 & 106 & 0.43 & 0\\
        Always retract & 3/3 & 12/12 & 170 & 173 & 0.41 & 0.42\\
        \textbf{Our method} & 3/3 & 12/12 & 87 & 157 & 0.35 & 0.17\\
    \end{tabular}
    \caption{Comparison to baseline strategies for trials in an environment with four elastic obstacles.}
    \label{tab:baselines}
    \vspace{-0.25in}
\end{table}

To test walking with and without our proposed method under the same conditions, We arranged four bungee cords on the anchor points of the obstacle rack shown in Fig.~\ref{fig:indoor setup} with heights from 12.7 cm to 22.9 cm high.
We compared the performance of our strategy to the unmodified gait as well as several modified open-loop baseline walking gaits, shown in Fig.~\ref{fig:baseline snapshots}.
Table~\ref{tab:baselines} summarizes the results comparing measurements of power and speed while walking freely ($P_f$ and $v_f$) and while in contact with obstacles ($P_o$ and $v_o$).

\subsubsection{Default baseline}
First we tested the default gait of Quad-SDK \cite{Norby2022}.
All parameters were left default (step height 0.07 m) except for the gait period which was matched to the other tests at 0.54s.
With default walking behavior, the first obstacle immediately halted the robot when it became stuck in the robot's front knees in all three trials.
\subsubsection{High stepping}
Tripling the swing-leg ground clearance from the default 0.07 m to 0.2 m entirely steps over some entanglements.
However, the outcome depends on the phase at which the leg strikes an obstacle; if the obstacles makes contact while the foot is rising, it can become jammed in the knee as in the default condition.
This strategy entangled a front leg on the first obstacle in the first trial and a back leg in the third, but successfully cleared the first obstacle only to get stuck on the third one in trial two.
Furthermore, high stepping incurs an energetic cost 50\% higher than the default even in the absence of obstacles.
\subsubsection{Knees forwards}
Modifying leg posture and hence the shape of the front surface of the leg appears a promising way to reduce entanglement.
Pointing the knees forwards (equivalent to walking backwards for Spirit 40 with its four symmetric legs) angles the lower shank down and back, allowing compliant obstacles beneath the knees to be naturally deflected down.
However, compliant obstacles above the knees of the leading legs become stuck on the robot's torso and those above the knees of the trailing legs become stuck at the hip joint.
This baseline crossed the first obstacle but became entangled in the third obstacle in all three trials.
Note that it is possible to angle the upper and lower shanks backwards by placing the feet far aft of the hips; however, this causes usual quadruped robots to tip over.
\subsubsection{Retraction on every step}
Retracting the leg according to the reaction motion on every step is capable of avoiding any entanglement that our proprioceptively-triggered reaction can avoid.
However, like the high stepping strategy, this behavior is energetically costly.
It crossed the obstacles in three out of three trials, but averaged about 170 W, double the usual power in the absence of obstacles.
\subsubsection{Our method}
Our proprioceptive detection and reaction strategy succeeded in all three out of three trials, overcoming obstacles that baselines 1--3 could not while using less power than aggressive baseline 4.
While our method was slower than baseline 4, it used as much power as the default in the absence of obstacles, half as much as baseline 4.

\subsection{Varying stiffness}
\label{sec:stiffnesses}

\begin{figure}
    \centering
    \includegraphics[width=1.0\linewidth]{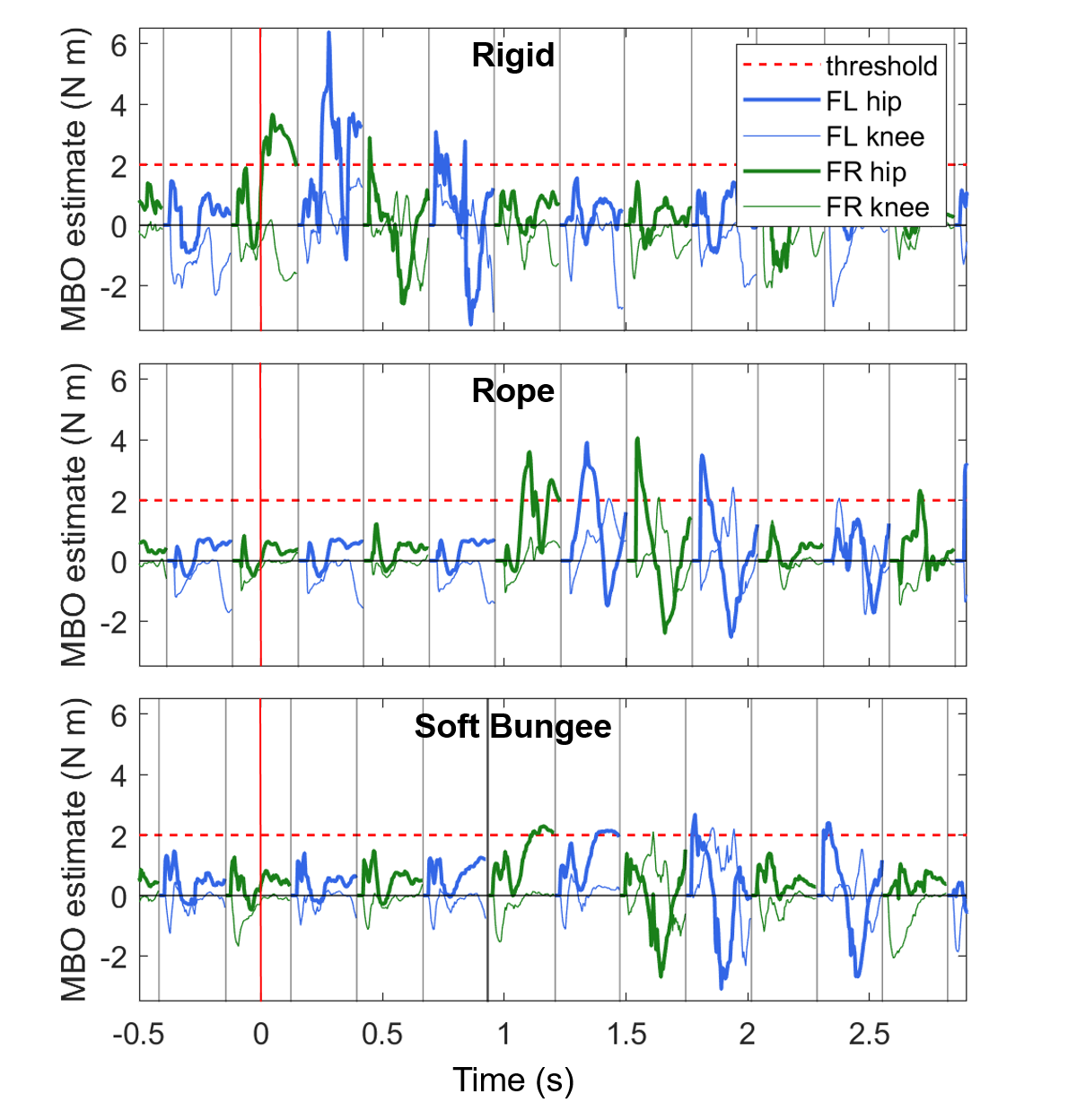}
    \caption{Momentum-based Observer (MBO) estimates for the front two legs as the robot walks into obstacles with various stiffnesses.
    Initial contact occurs at 0s, but the force is not significant enough to warrant reaction until slightly later for the soft bungee cord and stiff rope.}
    \label{fig:obstacles}
\end{figure}

\begin{table}
    \centering
    \begin{tabular}{l|lllll|l}
        Obstacle & Soft & Rope & Rigid & Mixed & Net & Total \\
        \hline
        \textbf{Our method} & 3/3 & 3/3 & 3/3 & 3/3 & 2/4 & 14/16
    \end{tabular}
    \caption{Overcoming obstacles of various stiffness}
    \label{tab:obstacles}
    \vspace{-0.25in}
\end{table}

Our proprioceptive entanglement reaction overcomes obstacles of various stiffnesses using identical control and estimation with no retuning for all experiments.
In addition to the arrangement of bungee cords, we tested our strategy against a softer bungee cord created by connecting three bungees in series, a stiff rope, and a short rigid aluminum beam.
The proprioceptive reaction strategy successfully detected and stepped over all four stiffnesses in all three out of three trials, summarized in Table~\ref{tab:obstacles}.

Since the detection and reaction strategy ignores small forces that can be pushed through, the reaction motion did not trigger until significantly later when contacting the soft bungee cord (see Fig.~\ref{fig:obstacles}).
By contrast, the reaction immediately triggered upon impact with the rigid beam.
For the stiff rope, the force remains negligible while the rope is slack but rapidly rises to be significant once the rope becomes taught.
This distinguishing between contact loads avoids overreacting to light contacts that are easily brushed aside while executing disentanglement motions once forces become large enough to impede forward progress.

\subsection{Many entanglements}
\label{sec:many entanglements}

\begin{figure}
    \centering
    \vspace{.5em}
    \includegraphics[width=1.0\linewidth]{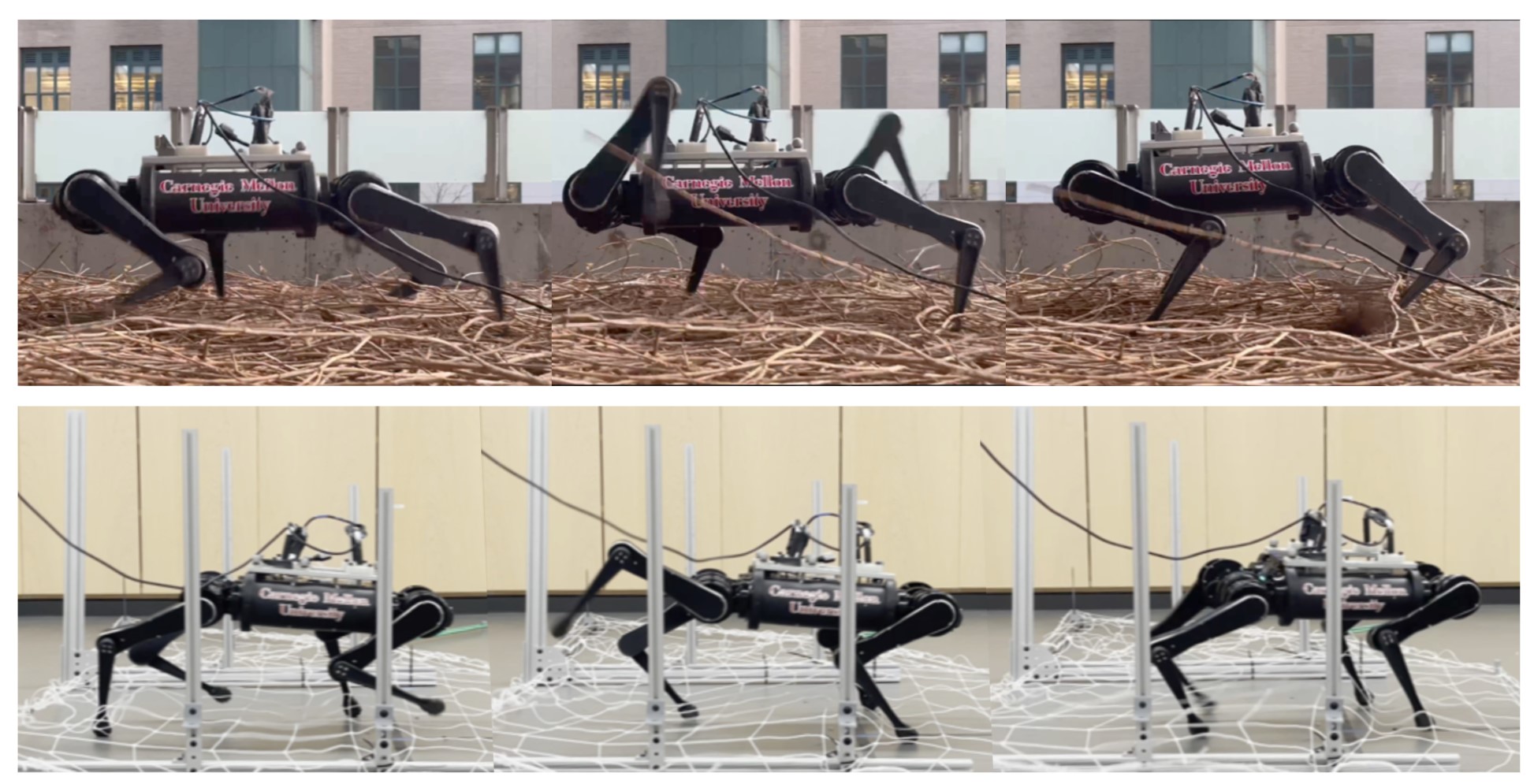}
    \caption{Quadruped robot walks through 8 cm of thick underbrush (Top) and a soccer net suspended 7.6 cm above the ground (Bottom)}
    \label{fig:challenge}
    \vspace{-0.15in}
\end{figure}

Subsequently, we tested our strategy in two challenging conditions.
Fig.~\ref{fig:intro} shows the robot walking through obstacles of several stiffnesses: rigid, bungees, and ropes shown in Fig.~\ref{fig:indoor setup}.
Our strategy again succeeded in all three out of three trials while interacting with obstacles of different stiffnesses simultaneously contacting different legs, demonstrating the proprioceptive reaction strategy's generality.

For an additional challenge, we walked the robot through a soccer net, presenting a great many small obstacles to the robot (Fig.~\ref{fig:challenge}).
For this extreme challenge, we increased the body height to 35 cm and decreased the commanded forward speed to 0.4 m/s.
The robot became entangled in the soccer net in two out of four trials, but succeeded in two.

\subsection{Outdoors}
\label{sec:outdoors}
Finally, to validate the viability of our momentum-based observer and underbrush controller, we ran a preliminary test of our system outside in a field of dense vines approximately 8 cm deep shown in Fig.~\ref{fig:challenge}.
We commanded Spirit 40  to walk forward at 0.4 m/s.
Our strategy advanced 1.65 m in 6.5 s while the standard controller's legs became immediately entangled and the robot advanced only 0.33 m before falling down.
Our strategy was significantly more robust and walked multiple gait cycles through thick underbrush before ultimately falling due to limitations in the camera-based body state estimation. Further outdoor experiments will require improved onboard body state estimation, and will be the subject of future work.

\section{CONCLUSIONS}

In this paper, we demonstrate a momentum-based observer for propriceptive joint torque estimation suited to classify contacts on a robot's limb as entanglements. In addition, we preset a simple leg reaction motion to disentangle from any detected entanglements.
These estimation and leg control modules can be easily integrated into an existing hierarchical planning and control framework to enable a commercial off-the-shelf quadruped robot to sense and disentangle from obstacles without hardware modification.
Hardware experiments demonstrate the efficacy of the proposed strategies and validate that the simplifying assumptions that enable decoupling are valid in physical realization.

Assumptions simplify our approach but limit the scope of our observer and reaction strategy.
The current implementation considers only forwards motion and while the extension strategy is spatial, the retraction strategy is not.
In particular, the retraction strategy will fail if it becomes stuck on something.
For cases such as this and for higher obstacles over which the robot would be forced to climb or leap, integration with higher-level planning will be required.

Future work can evaluate the performance of different walking, sensing, and reaction strategies in real environments from underbrush to thickets including performance metrics like the maximum speed that can be achieved among different numbers and types of obstacles.
Extraction of additional contact location and force magnitude estimates or data-driven estimation could provide refined obstacle information for disentanglement.
While the overall approach presented in this work should apply generally to platforms of other morphologies, extension to biped robots requires considerations of balance, foot shape, and other features.
This includes generalizing retraction to spatial motion including sideways walking in which the robot's hip and knee joint axes are parallel to the direction of travel.

Extensions could also address appropriate escalation of disentanglement maneuvers from simple behaviors that work for most situations to complex replanning to handle more circuitous extrication from particularly severe entrapment, as discussed in Section~\ref{sec:formulationandgait}.

\addtolength{\textheight}{-15cm}   





\bibliographystyle{IEEEtran}
\bibliography{references.bib}

\end{document}